\documentclass[runningheads]{llncs}
\usepackage{svg}
\usepackage{acro}
\usepackage{multirow}
\usepackage[T1]{fontenc}
\usepackage[caption=false]{subfig}
\usepackage{graphicx}

\DeclareAcronym{ICPR}{short = ICPR, long = International Conference on Pattern Recognition (ICPR)}
\DeclareAcronym{TSM}{short = TSM, long = Temporal Shift Module (TSM)}
\DeclareAcronym{HOG}{short = HOG, long = Histogram of Oriented Gradients (HOG)}
\DeclareAcronym{MHI}{short = MHI, long = Motion History Images (MHI)}
\DeclareAcronym{I3D}{short = I3D, long = Inflated 3D CNN (I3D)}
\DeclareAcronym{C3D}{short = C3D, long = Convolutional 3D (C3D)}

\begin{document}
\title{Action Recognition Using Temporal Shift Module and Ensemble Learning}

\author{Anh-Kiet Duong\inst{1}\thanks{The master program pursued by the first author is sponsored by the ``Vingroup Science and Technology Scholarship Program for Overseas Study for Master's and Doctoral Degrees''.}\orcidID{0009-0005-0230-6104} \\ \and
Petra Gomez-Krämer\inst{2}\orcidID{0000-0002-5515-7828}}
\authorrunning{A. K. Duong et al.}
\institute{University of Limoges, Limoges, France\\
\email{anh-kiet.duong@etu.unilim.fr}\\ \and
L3i Laboratory, La Rochelle University, France \\
\email{petra.gomez@univ-lr.fr}}
\maketitle              %
\begin{abstract}
This paper presents the first-rank solution for the Multi-Modal Action Recognition Challenge, part of the Multi-Modal Visual Pattern Recognition Workshop at the \acl{ICPR} 2024. The competition aimed to recognize human actions using a diverse dataset of 20 action classes, collected from multi-modal sources. The proposed approach is built upon the \acl{TSM}, a technique aimed at efficiently capturing temporal dynamics in video data, incorporating multiple data input types. Our strategy included transfer learning to leverage pre-trained models, followed by meticulous fine-tuning on the challenge's specific dataset to optimize performance for the 20 action classes. We carefully selected a backbone network to balance computational efficiency and recognition accuracy and further refined the model using an ensemble technique that integrates outputs from different modalities. This ensemble approach proved crucial in boosting the overall performance. Our solution achieved a perfect top-1 accuracy on the test set, demonstrating the effectiveness of the proposed approach in recognizing human actions across 20 classes. Our code is available online \footnote{https://github.com/ffyyytt/TSM-MMVPR}.

\end{abstract}

\section{Introduction}

Human action recognition is a computer vision task with the aim of recognizing and classifying human activities in images or videos. The objective is to analyze, understand human behavior and to classify the actions performed in the image or video into a predefined set of action classes. It is important for many applications such as video indexing, biometrics, surveillance and security. 
Human action recognition is a complex task due to the difficulty of extracting information about person’s identity and their psychological states \cite{ELHARROUSS2021103116}. Moreover, ambiguities in recognizing actions does not only come from the difficulty to define the motion of body parts, but also from many other challenges related to real world problems such as camera motion, dynamic background, and bad weather conditions \cite{JEGHAM2020200901}.

Human action recognition can be achieved using different sensors such as RBG or RGB-D cameras, but also depth sensors, infrared or thermal cameras. 2D RGB data have attracted huge attention as visual data are rich in features and less computational expensive than 3D data. Furthermore, RGB cameras are very common and affordable. However, combining modalities like depth or thermal can enhance recognition by providing complementary information, especially in challenging environments.

The Multi-Modal Action Recognition Challenge, part of the Multi-Modal Visual Pattern Recognition Workshop at \acl{ICPR} 2024, focuses on the complex task of recognizing human actions from multi-modal data sources. This track presents a unique opportunity to explore the integration of multiple data modalities—namely RGB, Depth, and thermal infrared (IR)  to improve the accuracy and robustness of action recognition systems. The challenge dataset comprises 2,500 videos, featuring 20 different action classes such as "shake hands," "ride a bike," "rope skipping," and potentially confusing pairs like "up the stairs" vs. "down the stairs," and "swivel" vs. "twist waist." With 2,000 videos designated for training and 500 for testing, each video spans a duration of 2 to 13 seconds and is captured in varying resolutions for different modalities, posing additional challenges for multi-modal fusion and feature extraction.

The objective of the competition is to advance the state-of-the-art in multi-modal action recognition by encouraging participants to develop innovative solutions that leverage the strengths of each modality while addressing the technical hurdles posed by feature heterogeneity and data fusion. To evaluate performance, the challenge uses the Top-1 and Top-5 accuracy metrics, which are commonly adopted benchmarks in the action recognition field. The organizers have provided a baseline model to guide participants, yet the competition aims to push beyond this baseline, setting new standards for multi-modal action recognition techniques.

This paper presents the 1st place solution for Track 3, which achieved a perfect score of 1.000 on the test set. Unlike typical multi-modal approaches, our solution focused solely on the thermal IR modality, as it yielded the highest accuracy during experimentation. We utilized the \acl{TSM} \cite{lin2019tsm} for efficient video understanding and applied transfer learning along with fine-tuning techniques specifically optimized for the thermal IR data. This targeted approach proved effective in overcoming the challenges of action recognition, even for closely related classes. By prioritizing a single modality, we simplified the model architecture while still surpassing the provided baseline and establishing a new benchmark for the challenge. The results highlight the potential of thermal IR data in this task's dataset, demonstrating its robustness and capability in accurately recognizing complex human actions.

The reminder of the article is organized as follows. We review related work in Section~\ref{sec:work}. Section~\ref{sec:method} presents our methods for action recognition based on TSM and ensemble learning. Section~\ref{sec:results} presents and discusses our results. Finally, we conclude our work in Section~\ref{sec:conclusion} and outlines future work.

\section{Related work}\label{sec:work}

We discuss below related work on video action recognition, deep learning architectures for video action recognition and ensemble learning.

\subsection{Video action recognition}
Video action recognition has been a key area of research in computer vision, evolving significantly with the advent of deep learning. Traditional approaches often relied on hand-crafted features, such as \acl{HOG} \cite{dalal2005histograms}, \acl{MHI} and optical flow, for capturing motion information, which were effective but limited by their inability to learn features directly from data in an end-to-end fashion \cite{wu2022survey}. However, these methods struggled to generalize across complex datasets due to their inability to learn data-driven representations.

With the introduction of deep learning, convolutional neural networks (CNNs) emerged as powerful tools for learning features directly from raw video data. Various architectures have been proposed, each with different strategies for capturing temporal and spatial information. These include:

\begin{itemize}
    \item 2D models: Methods like Temporal Segment Networks (TSN) \cite{wang2016temporal} extend 2D CNNs by applying temporal sampling and aggregation strategies across video frames, allowing them to capture long-range temporal dependencies. While the \acl{TSM} shifts part of the feature channels along the temporal dimension, enabling the model to capture temporal dependencies without the computational cost of full 3D convolutions.
    
    \item 3D models:  Approaches such as \acl{C3D} \cite{tran2015learning} and \acl{I3D} \cite{carreira2017quo} extend traditional 2D convolution to the temporal dimension, enabling joint spatial-temporal feature learning. These models have demonstrated superior performance on benchmark datasets but often require substantial computational resources.

    \item Two-Stream models: The two-stream architecture \cite{wu2022survey} consists of separate RGB and optical flow streams, which are later fused to combine spatial and motion information. Although effective, this approach relies heavily on the quality of optical flow estimation.
\end{itemize}

\subsection{Deep learning architectures}
Deep learning architectures have significantly advanced video action recognition. One of the foundational models is ResNet, introduced by He et al. \cite{he2016deep}. ResNet employs skip connections to facilitate the training of very deep networks, effectively addressing the vanishing gradient problem and enabling improved feature learning through residual learning. Its effectiveness in extracting rich features has made it a popular choice for various computer vision tasks.

Building on ResNet, ResNeXt \cite{xie2017aggregated} enhances the model capacity through a split-transform-merge strategy. This design allows ResNeXt to learn complex patterns more efficiently without a substantial increase in computational cost, making it particularly effective for high-accuracy tasks such as action recognition.

Another notable architecture is ResNeSt \cite{zhang2022resnest}, which extends ResNeXt by incorporating split-attention mechanisms. This enhancement improves the model’s ability to capture rich feature representations by dynamically focusing on different parts of the input data, further boosting performance in action recognition tasks.

Recent developments in transformer architectures, including the Vision Transformer (ViT) \cite{dosovitskiy2020image} and Swin Transformer \cite{liu2021swin}, have also shown promise in video analysis. ViT \cite{dosovitskiy2020image} treats video frames as sequences of patches, allowing it to capture long-range dependencies and global context more effectively than traditional CNNs. Meanwhile, the Swin Transformer \cite{liu2021swin} introduces a hierarchical architecture that operates at multiple scales, demonstrating strong performance across various benchmarks. Together, these transformer-based models highlight the potential for innovative approaches in enhancing action recognition tasks.

\subsection{Ensemble learning}
Ensemble learning has been widely used in action recognition to improve the accuracy and the robustness by combining predictions from multiple models. Techniques like Bagging, Stacking, and weighted averaging are popular, with more advanced methods such as AdaBoost and Gradient Boosting also explored \cite{mienye2022survey}. These approaches help to address individual model limitations by merging outputs from different modalities or architectures.

In this work, we chose weighted averaging for its simplicity and effectiveness, as well as time constraints, allowing us to efficiently combine predictions without adding excessive complexity.

\section{Action recognition method} \label{sec:method}
\begin{figure}[h]
    \centering
    \includegraphics[width=0.8\linewidth]{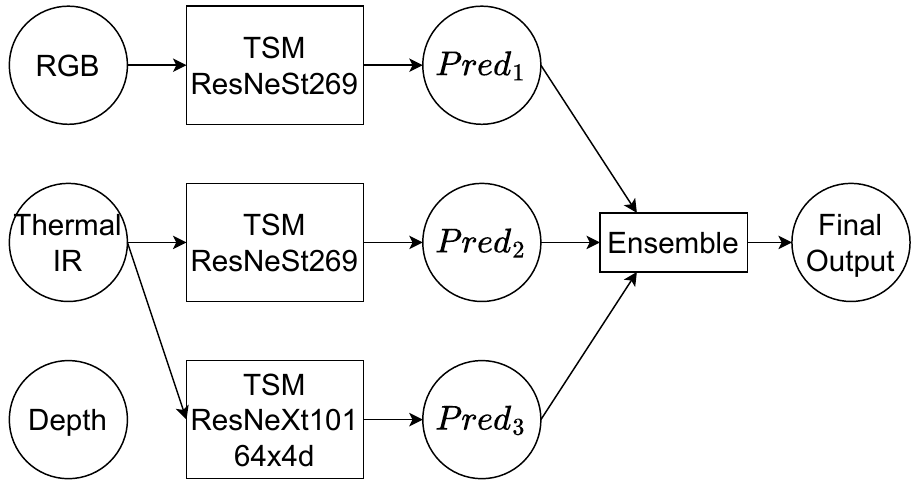}
    \caption{Overview of the prediction process in the proposed solution.}
    \label{fig:flow}
\end{figure}

In Figure \ref{fig:flow}, we illustrate the methodology employed in our solution for predicting human actions. Our method is centered around the \acl{TSM} \cite{lin2019tsm}. Specifically utilizing the ResNeSt-269 \cite{zhang2022resnest} variant, which is trained exclusively on RGB images, and also takes RGB data as input for prediction pharse. Additionally, we incorporate two TSM models trained on thermal IR images: one is the ResNeSt-269, and the other is the ResNeXt101 $64 \times 4d$ \cite{xie2017aggregated}, both of which also take thermal IR data as input. Notably, we opted not to use depth images in our approach due to their large size. We also tested other backbone models, such as ResNet \cite{he2016deep} and Inception \cite{szegedy2016rethinking}, but the results were not high on the leaderboard.
Therefore, we decided to use ResNeSt-269 and ResNeXt101 $64 \times 4d$ for our final solution since it provide higher score on the leaderboard. And we use both for ensemble learning because both have very high and almost perfect results on the test set.

The outputs from these three distinct models are subsequently ensembled to produce the final predictions. This strategy not only streamlines the computational resources required for training and inference but also enhances the overall performance of our system. By allowing each model to focus on a specific type of data input (RGB or thermal IR) we reduce the computational burden, as it eliminates the need to process all three data modalities simultaneously. This approach significantly decreases the resources needed for training and utilizing the models for prediction, enabling us to achieve more efficient results while maintaining high accuracy in action recognition.

\subsection{\acl{TSM}}

Following the baseline provided by the organizers, we continued to use the \ac{TSM} as the core of our solution for action recognition. To enhance performance, we explored different backbone models and input data modalities, as detailed in the following subsections.

\subsubsection{Backbone models}
Due to the public code provided by the authors of TSM \cite{lin2019tsm}, which is compatible with ResNet architectures, we experimented with various models, including ResNet \cite{he2016deep}, ResNeXt \cite{xie2017aggregated}, and ResNeSt \cite{zhang2022resnest}. Among these, ResNeXt and ResNeSt demonstrated superior performance compared to ResNet. Consequently, we selected ResNeXt and ResNeSt for the ensemble process, as they provided better accuracy and robustness for action recognition tasks.

\subsubsection{Data modality}
In our experiments, we utilized both RGB and thermal IR data. We opted not to explore depth data due to its substantial file size and the time constraints we faced during the challenge. Notably, our findings revealed that thermal IR images yielded better classification performance than RGB images. When using identical backbone architectures, models processing thermal IR input consistently outperformed those using RGB, highlighting the advantages of leveraging different data modalities for improved action recognition.

\subsection{Training Strategy} \label{sec:train}
Initially, we split the dataset into two distinct subsets: a training set and a validation set, consisting of 80\% and 20\% of the data, respectively. This partitioning results in 1,600 videos used for training and 400 videos reserved for validation. During the training phase, we train the models using the training set for $100$ epochs, while the validation set is employed to evaluate the models' performance and to select the most suitable hyperparameters. The validation process allows us to select the backbone model and fine-tune parameters such as learning rate, dropout rate, and number of segments,... thereby optimizing the models for the task of action recognition.

Given the limited size of the dataset, comprising only 2,000 videos in total, we take an additional step to improve model generalization. After identifying the optimal hyperparameters based on the validation set, we retrain the models using the entire dataset for $200$ epochs, including all 2,000 videos without the validation. This final training step aims to enhance the models' ability to learn from the complete set of available data, thereby maximizing their capacity to capture the underlying patterns within the action sequences.

By employing this two-step training strategy, we ensure that our models are well-calibrated and capable of achieving high performance despite the small dataset size. This approach allows us to make the most effective use of the available data, balancing the need for hyperparameter tuning with the advantages of training on the full dataset.

\subsection{Configuration}
As detailed in Section \ref{sec:train}, we selected our hyperparameters by training the models for a limited number of epochs on the validation set. This method allowed us to efficiently identify suitable configurations. In addition to the hyperparameters we fine-tuned, we retained several default settings from the strong public implementation of the \ac{TSM} \footnote{https://github.com/mit-han-lab/temporal-shift-module}, such as image size and augmentation techniques. Ultimately, the hyperparameters we converged on include an average type for ensemble weighting, a segment count of $8$, and a learning rate of $0.01$. These choices reflect our commitment to optimizing the model's performance.

\subsection{Ensemble learning}
In our solution, we employ a basic ensemble learning method as follows in \cite{jeon20201st} by combining the softmax outputs of each model, applying a specific weight $w_i$ to each model \( i \). This approach allows us to leverage the strengths of each individual model while mitigating their weaknesses, leading to improved overall performance. The final prediction $P$ is computed using the following formula:

\begin{equation}
P = \sum_{i=1}^{n} w_i \cdot Pred_i
\end{equation}

where $Pred_i$ represents the softmax output of model \( i \) and \( n \) denotes the total number of models in the ensemble. By adjusting the weights $w_i$, we can fine-tune the contribution of each model to the final prediction, thereby optimizing the ensemble's performance based on their respective accuracies and confidence levels.

This weighted ensemble strategy not only enhances the robustness of our predictions but also provides a means to effectively manage the trade-off between computational efficiency and predictive accuracy. Ultimately, this methodology enables us to achieve superior results in action recognition tasks, as the combined predictions from diverse models yield a more comprehensive understanding of the input data across different modalities.

Looking toward the future, our ensemble methodology could be expanded to incorporate more sophisticated ensemble techniques. For instance, algorithms such as AdaBoost \cite{freund1997decision}, which focuses on converting weak learners into a strong learner by adjusting weights iteratively, could be implemented to further enhance prediction accuracy. Additionally, more advanced techniques like Bagging and Stacking could also be explored \cite{mienye2022survey}. These ensemble methods have shown promise in various machine learning tasks and may provide further improvements in the context of multi-modal action recognition. By integrating these approaches, we aim to refine our ensemble strategy and explore new avenues for enhancing model performance in future research.

\section{Results} \label{sec:results}

In this section, we present our results on our validation sub-set and on the test set of the competition.

\subsection{Task description} The competition provided the participants with two datasets: a labeled training set consisting of 2,000 videos across 20 action classes, and an unlabeled test set of 500 videos. The goal was to predict the labels of the test videos, with each video allowing up to five ranked predictions for evaluation. The competition metrics used were Top-1 and Top-5 accuracy, based on the highest-ranked correct prediction and any correct prediction within the top five, respectively.

For model training, we initially split the training set into 80\% for training (1,600 videos) and 20\% for validation (400 videos) to optimize hyperparameters and evaluate performance. After this step, we retrained the models using the entire training set of 2,000 videos for 200 epochs to maximize the utilization of the available data. The dataset consisted of 20 action labels:

\begin{table}[h]
\begin{tabular}{|l|l|l|l|l|}
\hline
switch light         & up the stairs & pack backpack & ride a bike    & turn around       \\ \hline
fold clothes         & hug somebody  & long jump     & move the chair & open the umbrella \\ \hline
orchestra conducting & rope skipping & shake hands   & squat          & swivel            \\ \hline
tie shoes            & tie hair      & twist waist   & wear hat       & down the stairs   \\ \hline
\end{tabular}
\end{table}

\begin{figure}[h]
    \centering
    \subfloat[RGB image \label{fig:rgb}]{\includegraphics[width=0.6\linewidth]{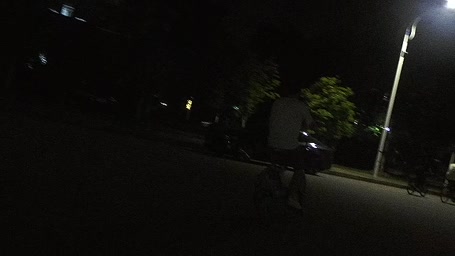}}
    \qquad
    \subfloat[Thermal IR image \label{fig:ir}]{\includegraphics[width=0.6\linewidth]{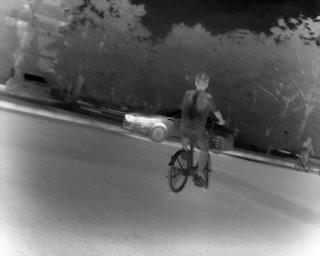}}
    \caption{Comparison of RGB and thermal IR images from the same frame of the "ride a bike" class.}
    \label{fig:rgb_ir}
\end{figure}

In Figure \ref{fig:rgb_ir}we show two samples taken from the same video and frame of the "ride a bike" class, but displayed in different formats: RGB and thermal IR. It is evident that the RGB image appears quite dark, making it difficult to discern details. In contrast, the thermal IR image clearly highlights the subject and action, offering better visibility. This observation helps explain why models with thermal IR input consistently outperformed those using RGB, as the IR modality provides more reliable information in low-light conditions, leading to better classification performance.

\subsection{Validation}
\begin{figure}
    \centering
    \includegraphics[width=0.75\linewidth]{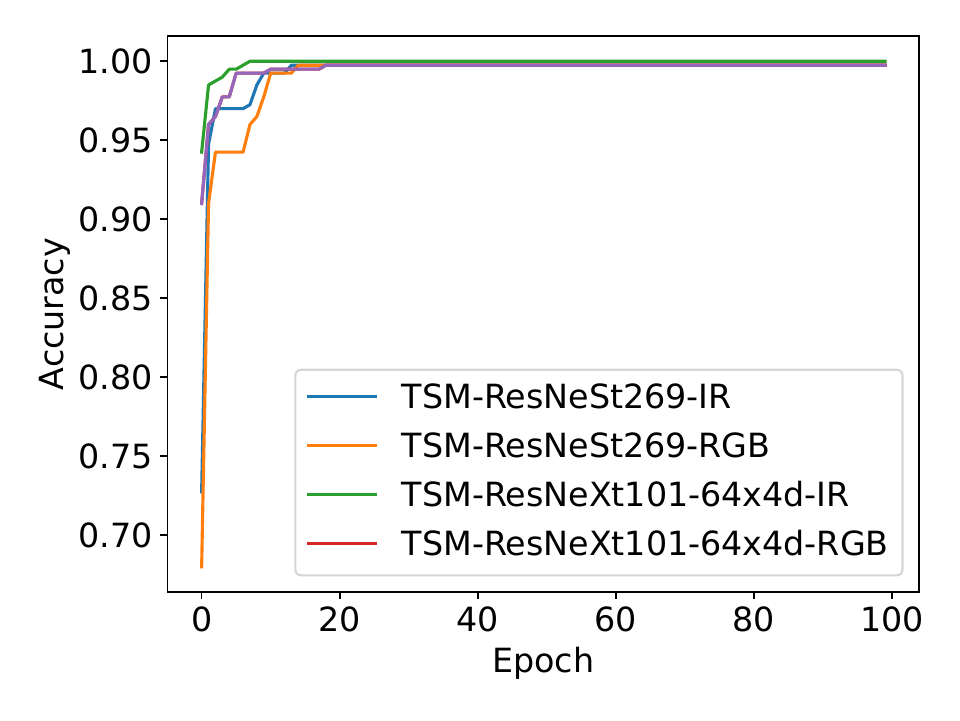}
    \caption{Validation accuracy of the ResNeSt269 and ResNeXt101 64x4d backbone models with thermal IR and RGB input on the validation set.}
    \label{fig:valid}
\end{figure}

In Figure \ref{fig:valid}, we present the validation accuracy results on 20\% of the dataset for validation, while training was conducted on the remaining 80\%. The figure illustrates the performance of the ResNeSt-269 and ResNeXt101 $64 \times 4d$ backbone models with thermal IR and RGB input, respectively. This step was primarily used to evaluate the models and determine suitable hyperparameters. After this preliminary validation phase, we retrained the models for $200$ epochs on the full set of 2,000 videos without a separate validation set, to make full use of the available data and maximize model performance.

\subsection{Test set}
Table \ref{tab:leaderboard} presents the results obtained on the test set of the competition. The results illustrate the effectiveness of different TSM configurations, backbones, and input modalities, highlighting the superior performance of the ensemble method. Notably, the TSM-ResNeSt-269 model stands out as the highest-performing single model, achieving a Top-1 accuracy of 0.9860 and a perfect Top-5 accuracy, showcasing its exceptional capability in action recognition tasks. However, it is important to note that ResNeSt-269 is also a backbone with very high computational complexity, which may pose challenges in terms of resource requirements during training and inference. This trade-off between performance and computational efficiency is a critical consideration in model selection for real-world applications.

The baseline model, which combines TSM with a ResNet50 backbone and incorporates all three input modalities (RGB, IR, and Depth), achieved an impressive Top-1 accuracy of 0.8643 and a Top-5 accuracy of 0.9940. It is particularly surprising that such high performance can be attained with the ResNet50 backbone, significantly outperforming configurations that rely solely on a RGB input. This result highlights the substantial benefits of integrating multiple data modalities for action recognition. However, the trade-off is that this approach demands significantly higher computational resources due to the need to process all three input types simultaneously.

Finally, our ensemble method achieved a perfect Top-1 accuracy of 1.0000 and a Top-5 accuracy of 1.0000, demonstrating the effectiveness of our approach in integrating multiple models to enhance overall performance. All the results show a clear trend: increasing computational complexity tends to improve model accuracy, whether by using larger backbone architectures or incorporating multiple input modalities. Based on this observation, we decided to maximize the size of our backbones while training each model on a single input type, and then combine them using ensemble learning. This strategy allowed us to achieve the highest performance without the need to train on all three input types simultaneously, thus optimizing both accuracy and computational efficiency.

\begin{table}[h]
\centering
\resizebox{0.6\linewidth}{!}{\begin{tabular}{lcc}
\hline
\multirow{2}{*}{Method}         & \multicolumn{2}{c}{Accuracy}                            \\ \cline{2-3} 
                                & Top-1                      & Top-5                      \\ \hline
TSM-ResNet50-RGB                & 0.5600                     & 0.9260                     \\
TSM-ResNeXt50-RGB               & 0.6660                     & 0.8760                     \\
Baseline                        & \multicolumn{1}{l}{0.8643} & \multicolumn{1}{l}{0.9940} \\
TSM-ResNeXt101-64$\times$4d-RGB & \multicolumn{1}{l}{0.8620} & \multicolumn{1}{l}{0.9700} \\
TSM-ResNeSt269-RGB              & 0.9620                     & 1.0000                     \\
TSM-ResNeXt101-64$\times$4d-IR  & 0.9820                     & 1.0000                     \\
TSM-ResNeSt269-IR               & 0.9860                     & 1.0000                     \\ \hline
Ensemble                        & \textbf{1.0000}            & 1.0000                    
\end{tabular}}
\caption{Performance comparison of various methods on the test set. }
\label{tab:leaderboard}
\end{table}

\section{Conclusion} \label{sec:conclusion}
In conclusion, our solution for the Multi-Modal Action Recognition Challenge at ICPR 2024 demonstrated that focusing on thermal IR data and leveraging the Temporal Shift Module (TSM) with an ensemble learning of selected models can achieve state-of-the-art results. The approach not only maximized accuracy but also highlighted the potential of using single-modality data in complex recognition tasks. Future research should explore advanced ensemble methods and additional modalities to further enhance performance and robustness in multi-modal action recognition systems.

\bibliographystyle{splncs04}
\bibliography{references}

\end{document}